\documentclass[conference]{IEEEtran}
\IEEEoverridecommandlockouts
\usepackage{cite}
\usepackage{amsmath,amssymb,amsfonts}
\usepackage{algorithmic}
\usepackage{graphicx}
\usepackage{textcomp}
\usepackage{xcolor}
\usepackage{multirow}
\usepackage{float}  
\usepackage{xspace}
\usepackage{tabularx}
\usepackage{url}
\usepackage{booktabs}
\usepackage{makecell}
\usepackage{siunitx}
\usepackage{diagbox}
\usepackage[caption=false,font=footnotesize]{subfig}
\newcommand{\hua}[1]{{\color{black}{#1}}}
\newcommand{\li}[1]{{\color{black}{#1}}}
\newcommand{\modelname}{\text{EMRFormer}\xspace}
\newcommand{\dcx}[1]{{\color{black}{#1}}}
\newcommand{\huarevising}[1]{{\color{black}{#1}}}

\def\BibTeX{{\rm B\kern-.05em{\sc i\kern-.025em b}\kern-.08em
    T\kern-.1667em\lower.7ex\hbox{E}\kern-.125emX}}
\begin{document}

\title{End-to-End Radar and Communication Modulation Recognition with Neuromorphic Computing\\
\thanks{\IEEEauthorrefmark{1}Corresponding author}
\thanks{$\dagger$ These authors contributed equally to this work}
\thanks{This work was supported by the National Key Research and Development Program of China (Grant No. 2024YDLN0008).}
}

\author{
\IEEEauthorblockN{
Xiaohu Li$^{a,\dagger}$, 
Chongxiao Qu$^{b,\dagger}$, 
Caiyong Lin$^{a}$, 
Chenxiao Dou$^{a,}$\IEEEauthorrefmark{1}, 
and Wei Hua$^{a}$}
\IEEEauthorblockA{$^{a}$
China Nanhu Academy of Electronics and Information Technology, Jiaxing, China}
\IEEEauthorblockA{$^{b}$
University of Electronic Science and Technology of China, Chengdu, China\\
\textit{Email:} lixiaohu@cnaeit.com, quchongxiao@cetc.com.cn, \{lincaiyong, douchenxiao, huawei\}@cnaeit.com}
}
\maketitle
\begin{abstract}
Although deep learning-based methods can achieve high accuracy in automatic modulation recognition (AMR) tasks, their high computational cost makes it difficult to strike a balance between accuracy and power consumption, thereby limiting their application on resource-constrained platforms.
Neuromorphic architectures that perform spike-driven inference with modest energy budgets have recently been explored for vision and time-series tasks. 
Motivated by these works, we propose \modelname, a novel end-to-end spiking nerural network (SNN) architecture that applies spike-driven transformer to the constraints of neuromorphic hardware for AMR.
The model incorporates an adaptive spike encoder and Integer Leaky Integrate-and-Fire neurons to \dcx{mitigate the degradation of effective information} and enhance SNN representational capacity.
By integrating spike-separable Convolution Neural Networks (SSCNN) into Spike-Driven Transformers (SpikeFormer), \modelname effectively extracts multi-scale temporal features from the raw IQ waveforms.
We validate our approach across various mainstream datasets,
the experimental results show that \modelname achieves state-of-the-art in terms of accuracy, outperforming all the baselines.
Furthermore, the model maintains strong performance in low signal-to-noise (SNR) environments and reduces theoretical energy consumption by over 90\%.
Finally, we evaluate our model on a KA200 neuromorphic chip. \li{The results show that our model achieves up to $ 5\times $ reduction in power compared to running on a 3090 GPU or an Orin NX.}
This work demonstrates a promising pathway for AMR on resource-constrained devices. 
\end{abstract}


\section{Introduction}
Automatic modulation recognition (AMR) is a critical technology in modern radar and communication systems.
It enables adaptive identification of modulation types under limited prior information, and is widely applied in electronic countermeasures, cognitive radio, and spectrum management~\huarevising{\cite{li2019survey,ER,lstm2,snn-LPI}}.

In recent years, many end-to-end deep learning approaches have achieved remarkable progress in AMR, such as convolutional-based neural networks \huarevising{~\cite{AWN,Conv-AMR}}, recurrent-based neural networks \huarevising{~\cite{lstm1,lstm2}}, and Transformer-based networks \huarevising{~\cite{amc-net,FEA-T}}, due to their powerful feature extraction and representation learning capabilities. 
These networks use the raw IQ signals as input, avoiding manual feature extraction or time-frequency pre-processing. 
The approaches significantly improve the accuracy while increasing its robustness across diverse AMR scenarios. 
However, these models often require a large number of learnable parameters, leading to high computational complexity~\cite{DL-AMR}.
Additionally, the dense multiply–accumulate (MAC) operations, which are the fundamental computation operations of these networks, need a large amount of computational resources, resulting in huge energy consumption and high latency.
These challenges hinder the deployment on resource-constrained platforms such as \huarevising{Unmanned Aerial Vehicles, IoT devices}, and mobile devices.



Brain-inspired neuromorphic computing techniques, which adopt the spiking neural network (SNN) as their core computational model, have attracted considerable attention for their biological plausibility, spike-driven processing, and potential for high energy efficiency \cite{roy2019towards,pei2019towards}.
In prior research works, SNNs have demonstrated competitive performance in computational vision \cite{spikeformer-v2,spikeyolo,spikeformer-v3}, speech \cite{speech}, physiological signal \cite{EEG-loihi}, and millimeter-wave radar tasks\cite{radar}. 
From a hardware perspective, running SNN-based networks on neuromorphic chips (e.g., Intel Loihi\cite{loihi}, IBM TrueNorth \cite{truenorth}, KA200\cite{KA200}) can greatly capitalize on the strengths of spike-driven and sparse accumulation operations, significantly reducing system-level energy consumption and latency\cite{dvs-ka200}.
\li{These advantages make AMR applications possible on resource-constrained platforms. 
However, existing SNN-based AMR methods\cite{sigmadelta,snn-LPI,SCNN3} typically rely on static spike encoder (e.g., Poisson encoder, temporal encoder) to convert signals into binary spikes, which inevitably leads to information loss. 
Meanwhile, these methods struggle with capturing long-term dependencies, failing to fully leverage temporal information. 
Although SNNs have a clear advantage in low power consumption on neuromorphic chips, their accuracy is not good enough in complex signal environments, making it difficult to effectively balance accuracy and power consumption.
This limitation constrains deployment in practical applications.
}

To address these issues, in this paper, we propose an end-to-end SNN-transformer-based architecture for the AMR task.
\li{We introduce an adaptive spike encoder that learns from IQ data to alleviate the information loss caused by static encoder. 
Meanwhile, the network uses Integer Leaky Integrate-and-Fire (ILIF) neurons, enhancing the representational capacity of SNNs while ensuring compatibility with neuromorphic chips during inference.}
Notably, inspired by the synergistic structure, we develop the  Spike-Separable Convolution Neural Network (SSCNN) and then integrate SSCNN into a spike-driven Transformer (SpikeFormer) {\hua{to capture the long-term dependencies from flattened multi-scale temporal dimensions while maintaining energy-efficient computation.}
We have conducted sufficient experiments to demonstrate that our method achieves State-of-the-Art performance among the end-to-end baselines.
It exhibits stronger robustness under low signal-to-noise Ratio (SNR) and cross-domain (radar/communication) conditions.
\li{Finally, the proposed model has been successfully validated on the neuromorphic chip KA200. Extensive experimental results demonstrate a power reduction of up to 5$×$ relative to conventional computing platforms, including the NVIDIA GeForce RTX 3090 GPU and NVIDIA Jetson Orin NX.}

The contributions of this work are summarized as follows:

\begin{enumerate}

\item  \huarevising{To the best of our knowledge, this is the first end-to-end SNN-based AMR framework that directly processes raw IQ signals to conduct a complete signal-to-category recognition.}
\item We develop the novel \modelname to effectively handle information loss by static spike encoder and capture long-term temporal dependencies in radar and communication signals.
\item We conduct the end-to-end training and testing on general-purpose platforms with model deployment to a KA200 chip for inference validation.
\end{enumerate}

\begin{figure*}[!t]
\centering
\includegraphics[width=\textwidth]{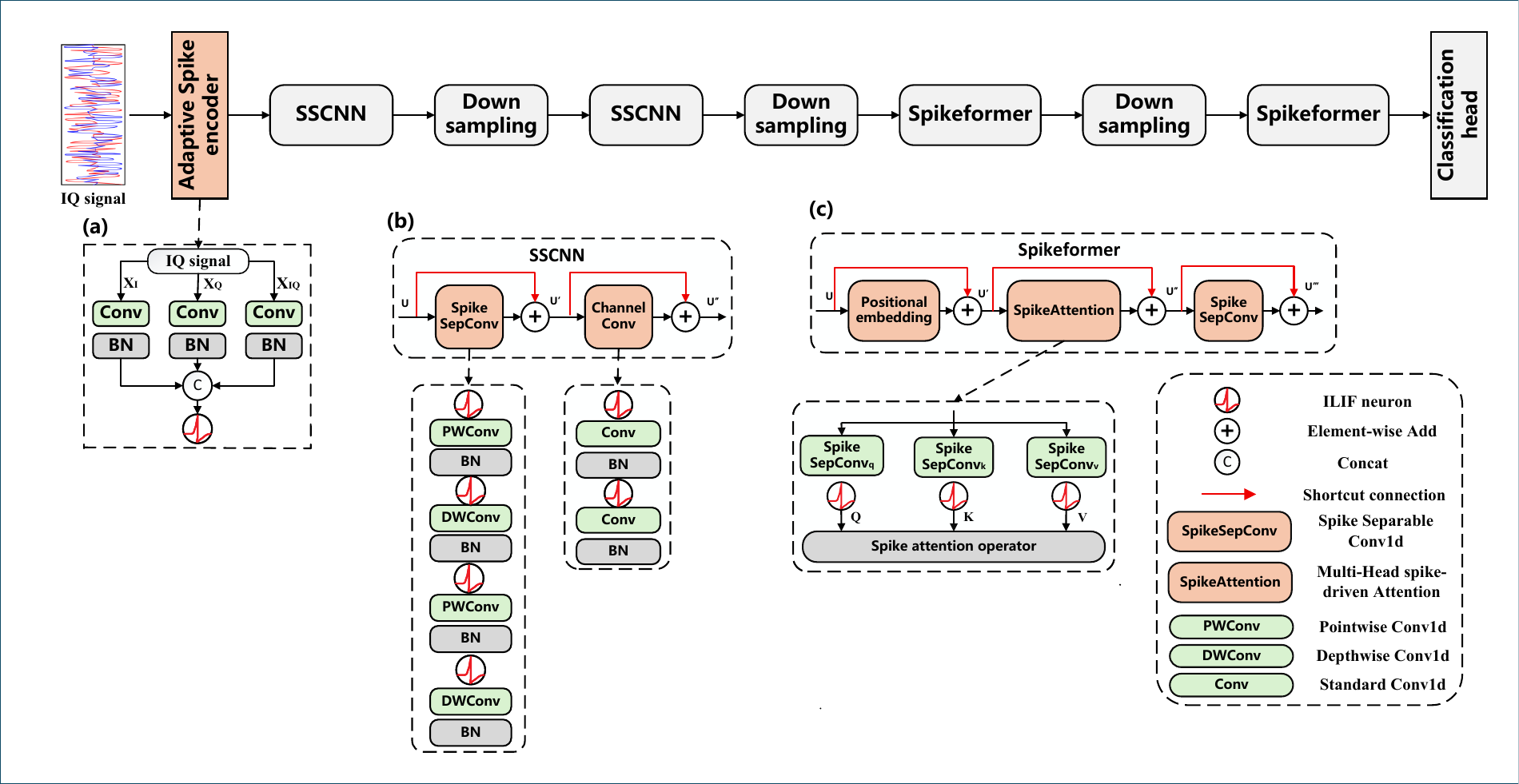}
\caption{Overall architecture of the \modelname.(a)Adaptive spike encoder Module: converts raw IQ to spikes train; (b)SSCNN Module: multi-scale local feature extraction; (c)SpikeFormer Module: long-term temporal modeling}
\label{EMR_SpikeFormer}
\end{figure*}
\section{Proposed Methodology}
In this section, we introduce the key components of \modelname step by step, including the ILIF neurons, an adaptive spike encoder, the SSCNN modules, and the SpikeFormer modules. The overall architecture of \modelname is illustrated in Fig. \ref{EMR_SpikeFormer}. 

\subsection{ILIF Neurons}
SNNs\cite{spikeformer-v2,spikeyolo,spikeformer-v3,speech,EEG-loihi,radar} are well-known for their energy efficiency. This feature renders SNNs particularly appealing for energy-intensive applications, such as electronic countermeasures\cite{ER} and \huarevising{cognitive radio\cite{li2019survey}}. 
However, the binary activation of SNNs results in the limitation of representational capacity, leading to the capture of subtle signal characteristics.

To resolve this problem, we adopt ILIF  (Integer Leaky Integrate-and-Fire) neuron \cite{spikeyolo}, which can implement the multi-spiking emission when SNN learning, to enhance the expressive power of SNNs in our \modelname.
Just as its name suggests, in an ILIF, the neuron can fire spikes up to $D$ times rather than only once, which means the neuron can transmit more information through the multi-fire mechanism throughout the networks. The formulation of an I-LIF neuron can be expressed as follows:
\begin{align}
U^{l}[t] &= H^{l}[t-1] + X^{l}[t], \label{eq:ilif1} \\
S^{l}[t] &= \text{Clip}\!\left(\text{round}(U^{l}[t]),\, 0,\, D\right), \label{eq:ilif2} \\
H^{l}[t] &= \beta \left(U^{l}[t] - S^{l}[t]\right), \label{eq:ilif3}
\end{align}
where $X^{l}[t]$ denotes the input current of the $l$-th layer at timestep $t$, $U^{l}[t]$ and $H^{l}[t]$ denote the membrane potentials before firing and after firing, respectively. 
$S^{l}[t]$ is the integer spike activation, and $\beta$ is the leakage factor.
$S^{l}[t]$ is an integer output, obtained by a rounding $U^{l}[t]$ and clipping it within the range $[0,D]$.
For examples, given $D=3$,  the corresponding activations $S^{l}[t]$ are $0$, $2$, and $3$ respectively,if $U^{l}[t]$ takes values $0.5$, $1.5$, and $4.5$. 
Compared to the regular LIF neurons that only trigger one spike, I-LIF neurons 
exhibit superior representational capacity without compromising the event-driven nature of SNNs.
In the rest of this paper,  we use $\mathrm{ILIF}(\cdot)$ as this $0$-$D$ activation function for convenience. 

\huarevising{Theoretically, the energy efficiency of neuromorphic chips is largely due to the binary activation in SNNs, as it requires on-chip computation to simpler accumulate (AC) operations rather than multiply–accumulate (MAC) operations.}
In practice, our proposed model converts the integer activations to binary activations via virtual time steps, tailored to work with the neuromorphic chips in the inference phase \cite{spikeyolo}.
 To implement this, the time step $T$ is extended to $T \times D$ steps,  $S^{l}[t]$ is rewritten as:
\begin{align}
\sum_{d=1}^{D} S^{l}[t,d] &= S^{l}[t], \qquad S^{l}[t,d]\in\{0,1\}.\label{eq:spike_sum}
\end{align}
While the input current $X$ of the $l+1$ layer is updated by Eq. \ref{eq:layer_input}:
\begin{align}
X^{l+1}[t] &=   \sum_{d=1}^{D}  W^{l} \cdot S^{l}[t,d], \label{eq:layer_input}
\end{align}
where $W^{l}$ is the matrix of synaptic weights.
$W^{l} \cdot S^{l}[t,d]$ performs AC operations as $S^{l}[t,d]$ is binary spikes.
Therefore, the inference process of the networks using ILIF demonstrates exceptional compatibility with the sparse computational capabilities of neuromorphic hardware.

\subsection{Adaptive Spike Encoder}
In most prior SNN-based AMR methods, the continuous-valued electromagnetic signal inputs are first converted into binary ($[0,1]$) representations before being fed into spiking neural networks. This additional data-format conversion introduces significant computational cost and incurs unavoidable information loss due to data degradation.
To fix this issue, we use the adaptive encoding ~\cite{adaptive-spike-encoding} in the encoder of our model. 
Firstly, we define $X_{I}\in \mathbb{R}^{1\times N}$ and $X_{Q}\in \mathbb{R}^{1\times N}$ as the in-phase and quadrature components to simplify the representation, respectively.
Thus, the raw data of IQ waveform is formalized into $X_{IQ} = [X_I, X_Q] \in \mathbb{R}^{2 \times N}$, where $N$ is the signal length. After that, $X_{I}$, $X_{Q}$, and $X_{IQ}$ are forwarded through three CNNs to perform feature extraction, shown in Fig.~\ref{EMR_SpikeFormer}.
The three outputs of CNNs are fused by a concatenation, and then are fed into an ILIF neuron. The processing is as follows:
\begin{align}
U &= \text{Concat}\big(
      \text{BN}(\text{Conv}(X_{I})),
      \text{BN}(\text{Conv}(X_{Q})), \nonumber\\
  &\qquad\quad
      \text{BN}(\text{Conv}(X_{IQ}))
    \big),\\
S &= \mathrm{ILIF}\big(U),
\label{eq:sn_ilif}
\end{align}
where $BN(\cdot)$ is the Batch Normalization, and the $ILIF(\cdot)$ emits at most $D$ spikes; the output $S$ is thus encoded into an integer number within $[0,D]$. 

As a learned model, the encoder network encodes the input signals in a data-dependent fashion, mitigating information loss caused by the prior biases inherent in static encoding. Moreover, eliminating the additional data-preprocessing step significantly reduces computational and storage demands on hardware resources.
Its computational efficiency is particularly advantageous for resource-constrained edge devices based on neuromorphic hardware.
\subsection{SSCNN Module}
AMR signal typically exhibits pronounced spatiotemporal characteristics, which necessitate the use of wider and deeper neural networks for feature extraction.
Following the design of Separable Convolution\cite{SepConv}, whose separable pointwise and depthwise convolution effectively facilitates the increase of networks' depth or width, we introduce an efficient spike-driven convolution module, named as Spike Separable Convolutional Neural Network (SSCNN).
As shown in Fig~\ref{EMR_SpikeFormer}, an SSCNN module works as eq. \ref{eq:sscnn1} and eq. \ref{eq:sscnn2}:
\begin{align}
U' &= U + \text{SpikeSepConv}(U), \label{eq:sscnn1} \\
U'' &= U' + \text{ChannelConv}(U'), \label{eq:sscnn2} 
\end{align}
where $U$ is the layer input.
$SpikeSepConv(\cdot)$ employs a stacked structure of pointwise and depthwise convolution to capture local signal features:
\begin{equation}\label{eq:spikesepconv}
\begin{split}
\text{SpikeSepConv}(U) &= \text{Conv}_{\text{dw}}\big(\mathrm{ILIF}\big(\text{Conv}_{\text{pw}}\big(\mathrm{ILIF}\big(\\
&\hspace{-1.8em}\text{Conv}_{\text{dw}}\big(\mathrm{ILIF}\big(\text{Conv}_{\text{pw}}(\mathrm{ILIF}\big(U))\big)\big)\big)\big)\big),
\end{split}
\end{equation}
where $Conv_{dw}(\cdot)$ and $Conv_{pw}(\cdot)$ denote one-dimensional depthwise and pointwise convolutions, respectively. $ChannelConv(\cdot)$ performs feature fusion across channels with nonlinear spike activations:
\begin{align}
\text{ChannelConv}(U') &= \text{Conv}(\mathrm{ILIF}\big(\text{Conv}(\mathrm{ILIF}\big(U')))).
\label{eq:channelconv}
\end{align}

As is widely known, the short connection is an effective technique in ANNs to resolve the degradation problem in deep networks \cite{spikeformer-v2,spikeformer-v3}.
However, in SNNs, the usage of short connections needs to not only adhere to the principle of identity mapping but also follow the paradigm of spike-driven paradigm.
As a result, we first route the input membrane to the later layer along the shortcut path. 
Skipping over one or more layers, the membrane directly performs element-wise operations with the output of the stacked layers.
Then, the result is passed into an ILIF neuron to trigger spikes. 
In this way, we implement a spike-driven version of a short connection, which greatly relieves the degradation problem in deep SNNs.
\subsection{SpikeFormer Module}
While SSCNN module gives an efficient method to extract spatiotemporal features, it falls short in modeling long-range dependencies between signals, which are deemed as an important factor to AMR task\cite{amc-net}.
Enlightened by the Transformer model, whose attention mechanism has a great success in evaluating relationships between widely separated tokens, we 
adopt a spike version of transformer\cite{spikeformer-v2} named SpikeFormer~\cite{spikeformer-v2} in our model to capture the long-range dependency.

The workflow of SpikeFormer is depicted in Fig.~\ref{EMR_SpikeFormer}(c), and the corresponding computations are as follows:
\begin{align}
U' &= U + \text{PosEnc}(U), \label{eq:spikeformer1} \\
U'' &= U' + \text{SpikeAttention}(U'), \\
U''' &= U'' + \text{SpikeSepConv}(U'').
\end{align}
where $\text{PosEnc}(\cdot)$ is the function of sinusoidal positional encoding.  $\text{SpikeAttention}(\cdot)$ denotes Multi-Head spike-driven attention, which computes spike-driven Q, K, and V via $SpikeSepConv$, and calculates the attention scores using the spike attention operator:
\begin{align}
Q &= \mathrm{ILIF}\big(\text{SpikeSepConv}_q(U')\big), \label{eq:qs} \\
K &= \mathrm{ILIF}\big(\text{SpikeSepConv}_k(U')\big), \label{eq:ks} \\
V &= \mathrm{ILIF}\big(\text{SpikeSepConv}_v(U')\big), \label{eq:vs} \\
\text{SpikeAttention}(U') &= \mathrm{ILIF}\big((QK^{T})V)).
\end{align}

The self-attention mechanism of SpikeFormer complements SSCNN by modeling dependency features across long-distance signals.
This enables \modelname to learn both short- and long-range dependencies from raw IQ sequences in an event-driven and energy-efficient manner.
In a short summary, capitalizing on the highly sparse network structures, SSCNN and SpikeFormer modules bring significant savings on computation and memory, making them deployment-friendly to neuromorphic chips.
With the twofold performance gains from SNN model and the neuromorphic chip, our proposed method is both practical and competitive to apply on AMR. 

\section{Experimental Results}
This section introduces the details of the experiment, including data collection, implementation,  evaluation methods, and ablation studies.

\subsection{Dataset and Experiment Settings}
\textbf{Datasets.} We evaluate \modelname  on three communication modulation datasets including RML2016.10a~\cite{rml2016a}, RML2016.10b~\cite{rml2016b}, RML2018.01a~\cite{rml2018}, and one radar modulation recognition dataset DeepRadar2022~\cite{lstm2}. Notably, DeepRadar2022 encompasses 23 distinct radar modulation types, while RML2018.01a includes 24 modulation types.  All samples in these two datasets are represented as 2$\times 1024$ IQ matrices. In contrast, RML2016.10a and RML2016.10b, which are generated using GNU Radio, include 11 and 10 modulation types, respectively, using samples formatted as 2$\times$128 IQ matrices.

\textbf{Experimental settings.} All experimental models are implemented in PyTorch and trained on an NVIDIA A100 GPU. The training process utilized the AdamW optimizer with an initial learning rate of 0.001 and a cosine annealing strategy for dynamic adjustment. Further training parameters include a batch size of 1024, time step T=1, integer approximation $D=4$, and a total of 150 epochs. The communication datasets (RML2016.10a, RML2016.10b, and RML2018.01a) are split in ratios of 6:2:2 for training/validation/testing purposes, while DeepRadar2022 employed its official split~\cite{lstm2}. 

To accommodate the varying sequence lengths across datasets, we apply different architectures to test the \modelname: on short-sequence datasets (RML2016.10a and RML2016.10b), our model utilizes 1 SSCNN layers and 2 SpikeFormer layers, whereas when testing on long-sequence datasets (RML2018.01a and DeepRadar2022), the \modelname employs 4 SSCNN layers and 2 SpikeFormer layers.

\textbf{ Baselines and Evaluation Metrics.} We compare our \modelname with four  end-to-end AMR SOTA methods, including AMC-NET~\cite{amc-net}, LSTM~\cite{lstm2}, FEA-T~\cite{FEA-T}, and AWN~\cite{AWN}. 
\hua{In line with prior works, we employ the overall accuracy (OA) across all SNR ranges to evaluate the classification performance and theoretical energy consumption to conduct energy analysis.}

\textbf{ Energy Consumption Calculation}. For a 1-D convolution layer, the energy consumption of ANNs and SNNs is calculated as follows:
\begin{align}
E_{\text{ANN}} &= O \times C_{\text{in}} \times C_{\text{out}} \times K \times E_{\text{MAC}}, \label{eq:ann_energy} \\
E_{\text{SNN}} &= (T \times D) \times fr \times O \times C_{\text{in}} \times C_{\text{out}} \times K \times E_{\text{AC}}, \label{eq:snn_energy}
\end{align}
where $O$ is the size of the output feature. $C_{\text{in}}$ and $C_{\text{out}}$ represent the size of the input and output channels, respectively. $K$ is the kernel size, $fr$ is the firing rate of spiking neurons. 
We assume that all the operations are executed on 
 a 32-bit floating-point, 45\,nm hardware for implementation, where $E_{\text{MAC}}=4.6\,\text{pJ}$ and $E_{\text{AC}}=0.9\,\text{pJ}$\cite{power, spikeformer-v2,spikeformer-v3}.
\begin{table*}[]
\centering
\caption{Performance Comparison of Baseline Models and \modelname Across Datasets}
\label{tab:performance}
\begin{tabular}{@{}cccccccccccccc@{}}
\toprule
\multirow{2}{*}{\diagbox[width=2.2cm,height=3.6em,trim=l]{Models}{Datasets}} & \multicolumn{1}{c}{
  \multirow{2}{*}{
    \parbox[c][3em][c]{0.5cm}{\centering input}
  }
} & \multicolumn{3}{c}{RML2016.10A}             & \multicolumn{3}{c}{RML2016.10B}                                                                          & \multicolumn{3}{c}{RML2018.01a}                                                                          & \multicolumn{3}{c}{DeepRadar2022}                                                                       \\ \cmidrule(l){3-14} 
                  & \multicolumn{1}{c}{}                       & \begin{tabular}[c]{@{}c@{}}Param\\ (M)\end{tabular} & \begin{tabular}[c]{@{}c@{}}Energy\\ (pJ)\end{tabular} & OA    & \begin{tabular}[c]{@{}c@{}}Param\\ (M)\end{tabular} & \begin{tabular}[c]{@{}c@{}}Energy\\ (pJ)\end{tabular} & OA    & \begin{tabular}[c]{@{}c@{}}Param\\ (M)\end{tabular} & \begin{tabular}[c]{@{}c@{}}Energy\\ (pJ)\end{tabular} & OA    & \begin{tabular}[c]{@{}c@{}}Param\\ (M)\end{tabular} & \begin{tabular}[c]{@{}c@{}}Energy\\ (pJ)\end{tabular} & OA    \\ \midrule
AWN & IQ & 0.12 & 2.76×10\textsuperscript{7} & 61.56 & 0.12 & 2.76×10\textsuperscript{7} & 64.23 & 0.33 & 3.06×10\textsuperscript{8} & 62.55 & 0.33 & 3.06×10\textsuperscript{8} & 78.54 \\
AMC-NET & IQ & 0.47 & 7.85×10\textsuperscript{7} & 61.15 & 0.47 & 7.85×10\textsuperscript{7} & 64.63 & 4.61 & 6.36×10\textsuperscript{8} & 58.36 & 4.61 & 6.36×10\textsuperscript{8} & 75.39 \\
FEA-T & IQ & 0.17 & 3.67×10\textsuperscript{7} & 59.69 & 0.17 & 3.67×10\textsuperscript{7} & 63.83 & 0.34 & 1.17×10\textsuperscript{8} & 55.82 & 0.34 & 1.17×10\textsuperscript{8} & 80.22 \\
LSTM & IQ & 0.3  & 1.18×10\textsuperscript{8} & 58.51 & 0.3  & 1.18×10\textsuperscript{8} & 64.44 & 0.3  & 9.41×10\textsuperscript{7} & 61.2  & 0.3  & 9.41×10\textsuperscript{7} & 85.73 \\
\modelname & IQ & 0.31 & \textbf{2.61×10\textsuperscript{6}} & \textbf{62.50} & 0.31 & \textbf{3.02×10\textsuperscript{6}} & \textbf{64.84} & 0.53 & \textbf{2.59×10\textsuperscript{7}} & \textbf{62.85} & 0.53 & \textbf{2.45×10\textsuperscript{7}} & \textbf{87.40} \\ 
\bottomrule
\end{tabular}
\end{table*}
\subsection{Performance Comparison}
Table \ref{tab:performance} presents the evaluation metrics for the aforementioned datasets. \modelname demonstrates superior performance, achieving high accuracy while reducing energy consumption by over 90\%.  On RML2016.10A, RML2016.10B, and RML2018, it attains average accuracies of 62.50\%, 64.84\%, and 62.85\%, respectively, surpassing CNN- and RNN-based baselines. On DeepRadar2022, \modelname achieves the accuracy of 87.4\%, a 2–10\% improvement over other baseline models, highlighting its cross-domain generalization ability between radar and communication domains. Figure \ref{fig:snr-accuracy} depicts the recognition accuracy across different SNR levels. \modelname maintains an accuracy edge in low-SNR conditions ($\textless$0 dB), showcasing robust feature extraction under heavy noise. In high-SNR scenarios ($\geq$0 dB), its accuracy curve rapidly saturates and remains superior. 

\huarevising{Fig. \ref{fig:acc_power} shows the comparison for the different models in terms of accuracy, power, and number of parameters. 
The experimental results demonstrate the superior performance of \modelname, surpassing prior works while maintaining high accuracy and significantly reducing power consumption. It achieves a well-balanced trade-off between performance and energy efficiency. 
These findings highlight the advantages of \modelname in terms of performance, robustness, and energy efficiency, positioning it as a promising candidate for low-power, real-time neuromorphic modulation recognition applications.}
\begin{figure}[!t]
  \centering
  \subfloat[RML2016.10a]{%
    \includegraphics[width=0.24\textwidth]{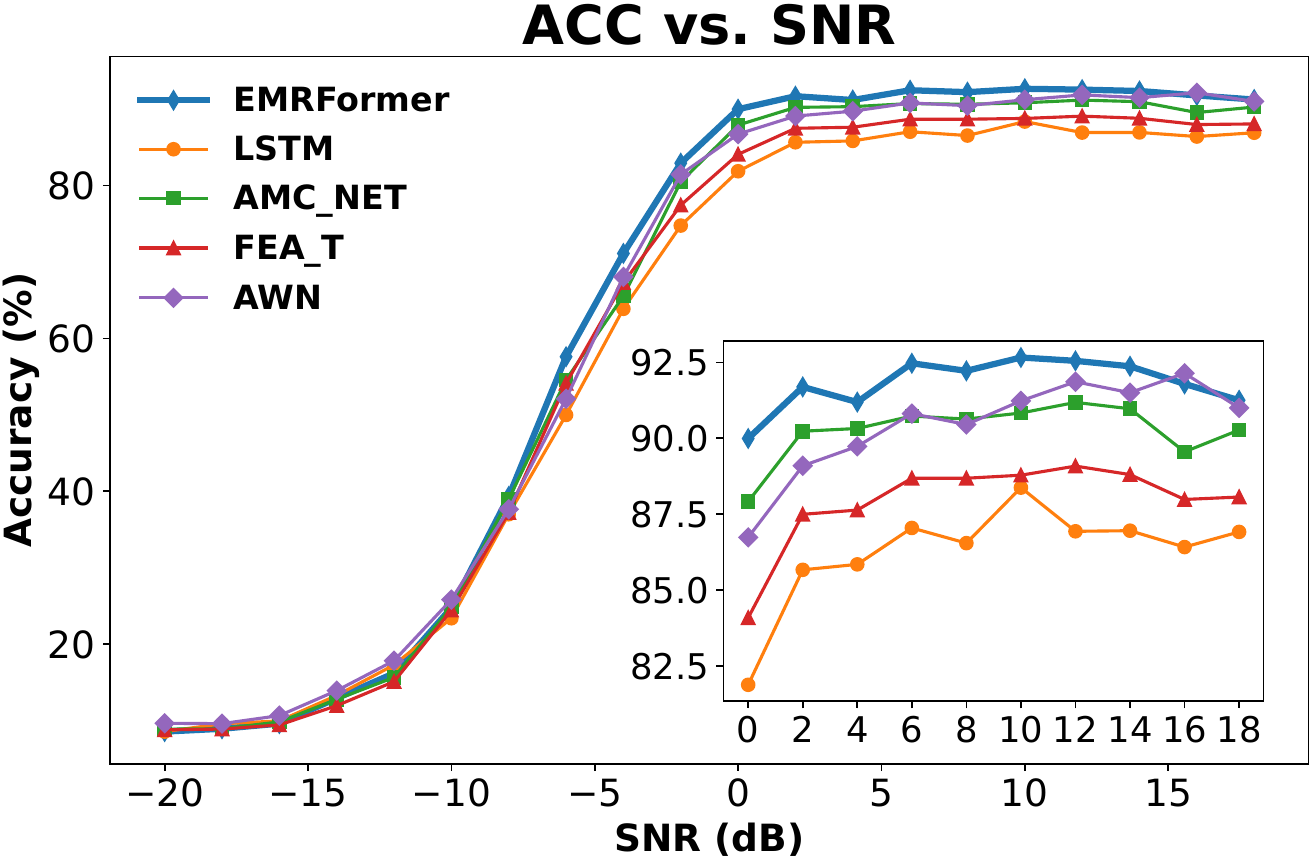}}%
  \hfil
  \subfloat[RML2016.10b]{%
    \includegraphics[width=0.24\textwidth]{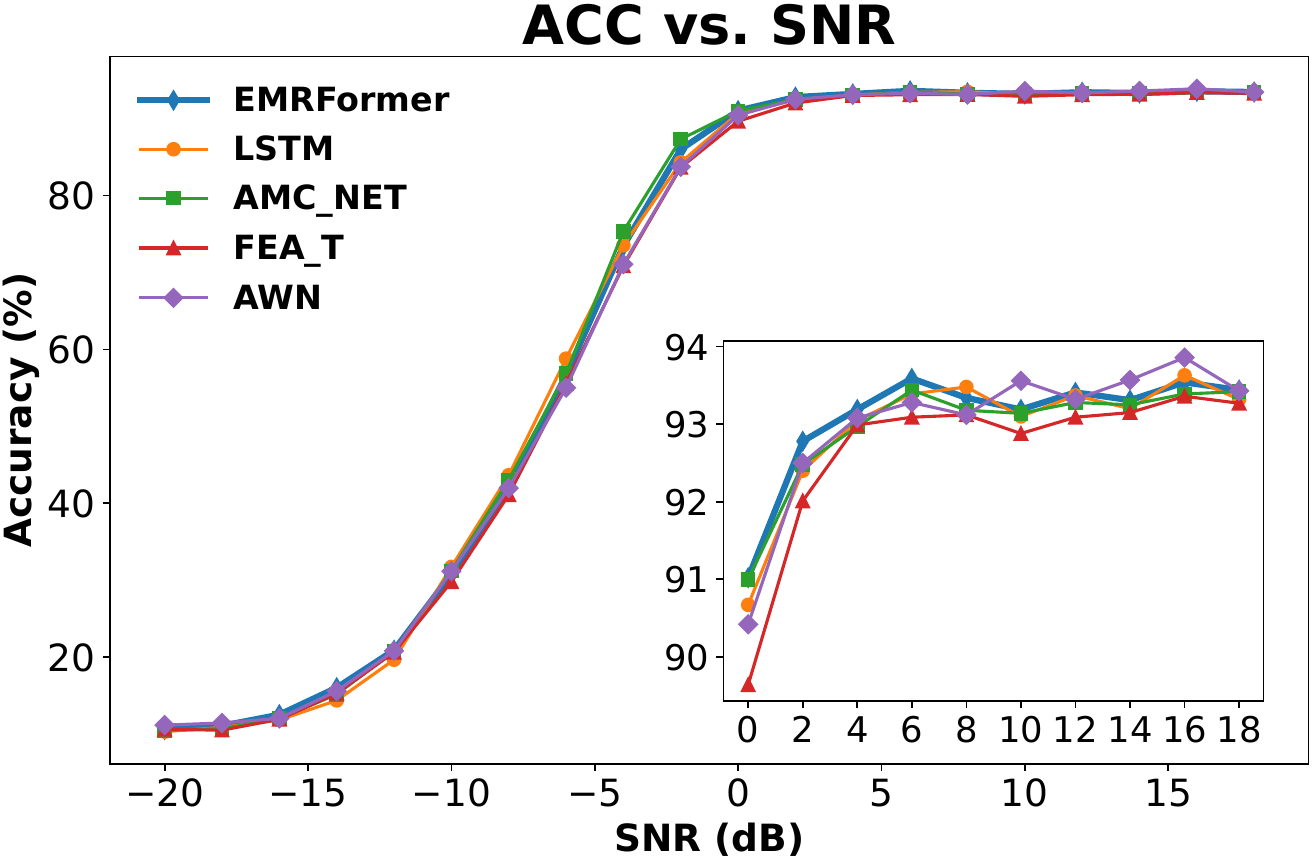}}%
   \hfil
  \subfloat[RML2018]{%
    \includegraphics[width=0.24\textwidth]{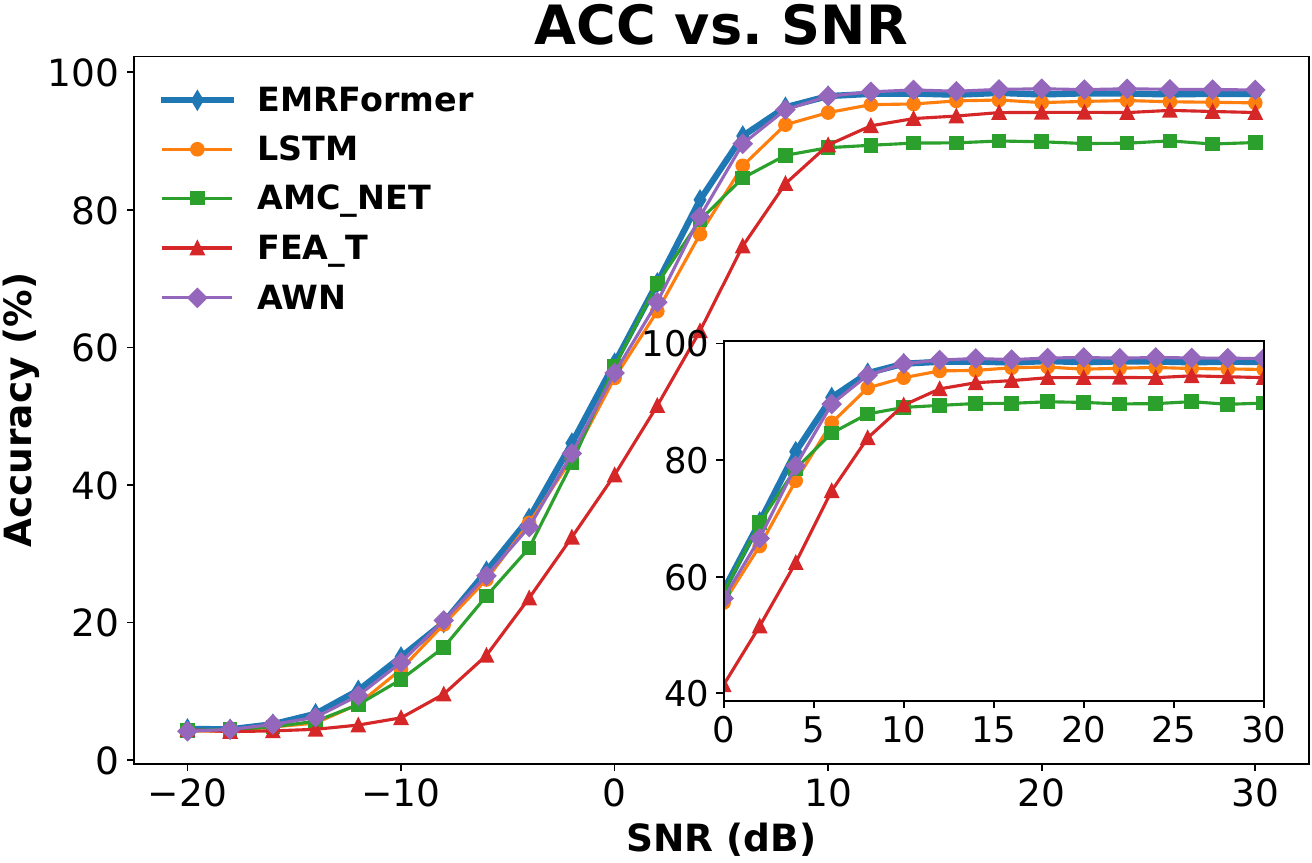}}%
  \hfil
  \subfloat[DeepRadar]{%
    \includegraphics[width=0.24\textwidth]{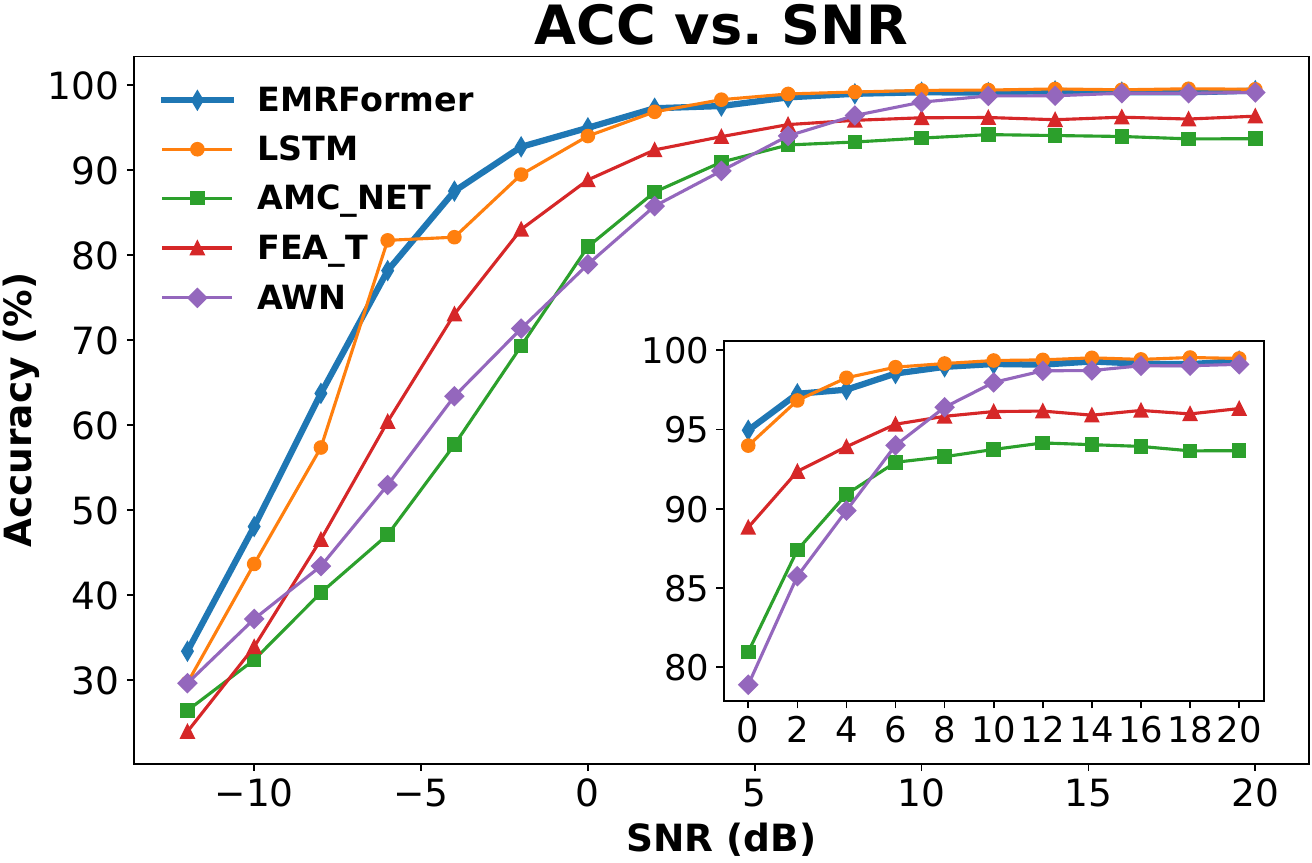}}%
  \caption{Accuracy-SNR curves on different datasets.}
  \label{fig:snr-accuracy}
\end{figure}
\begin{figure}[!t]
\centering
\includegraphics[width=0.4\textwidth]{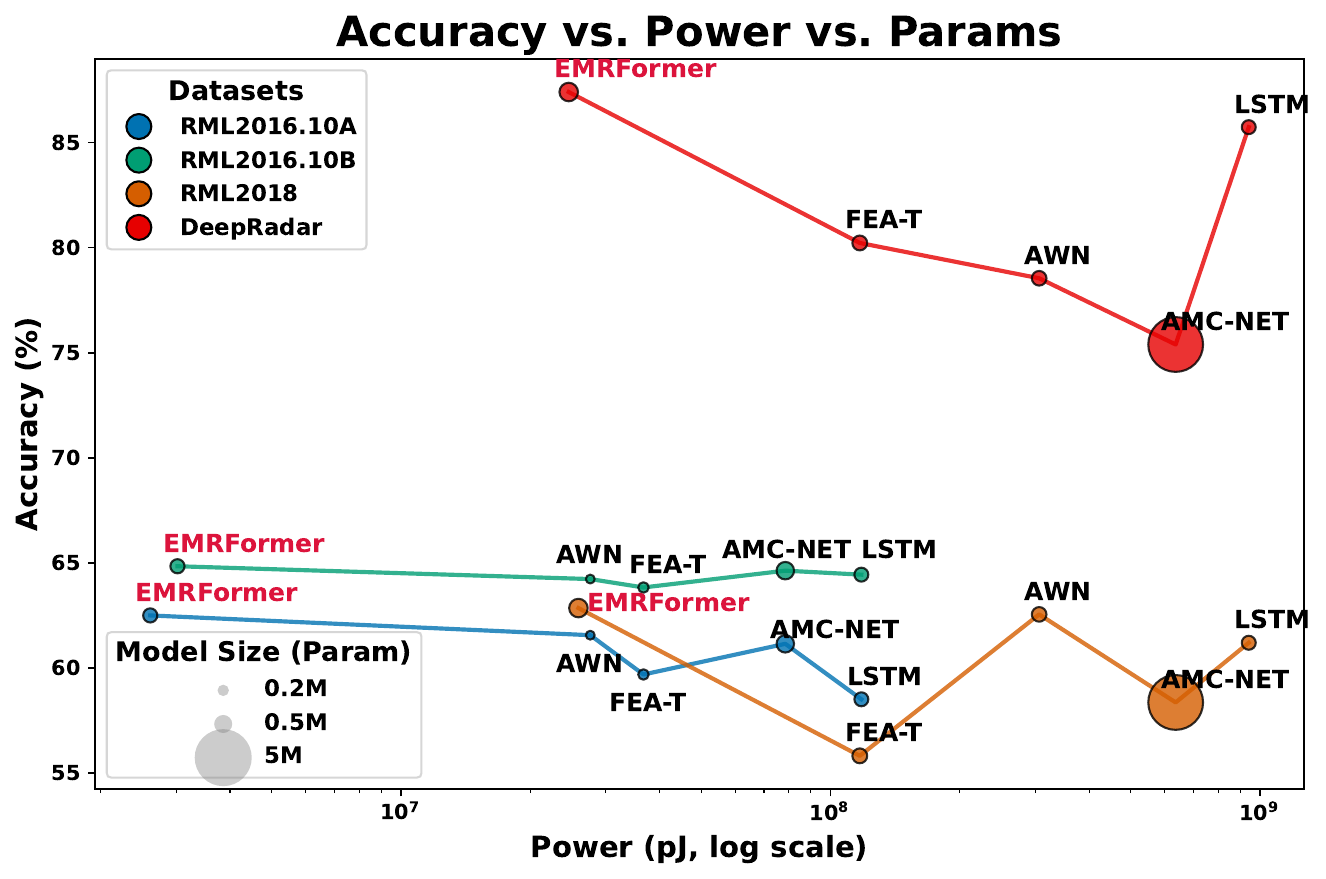}
\caption{\modelname versus baseline methods on four datasets}
\label{fig:acc_power}
\end{figure}
\subsection{Ablation Study}
The ablation study on \modelname using the Deepradar dataset reveals that altering any component leads to a performance drop, as summarized in \ref{tab:ablation}. Switching Conv1d to Conv2d boosts parameters and energy, while it cuts the performance of accuracy. 
Axing PosEnc further hits accuracy, and swapping I-LIF with LIF can reduce energy consumption, whereas it severely worsens results. 
Swapping the Adaptive spike encoder for static encoder (poisson encoder) slashes accuracy and hikes power draw. 
Figure \ref{fig:ablation} indicates that variants match the full model with the high SNR, but they fall behind at low SNR. The full model, packing only $0.53 M$ parameters, surpasses all altered versions, proving its design excels at sparse feature extraction, temporal modeling, and low-SNR resilience.
\begin{table}[]
\caption{The ablation results of the \modelname on the DeepRadar}
\label{tab:ablation}
\begin{tabular}{lccc}
\hline
Model                      & Param(M) & Energy(pJ) & ACC(\%)  \\ \hline
\modelname           & 0.53M       & $2.45\times10^{7}$   & 87.40         \\
Conv1d→Conv2d              & 1.25M    & $7.20\times10^{7}$     & 86.34 (-1.06)  \\
Without PosEnc     & 0.53M   & $2.64\times10^{7}$      & 86.05 (-1.35)  \\
I-LIF→LIF                  & 0.53M   & $2.45\times10^{7}$      & 76.71 (-10.69) \\ 
\begin{tabular}[c]{@{}l@{}}Adaptive spike encoder→\\ Poisson encoder\end{tabular} & 0.53M   & $1.96\times10^{8}$      & 79.32 (-8.08) \\ \hline
\end{tabular}
\end{table}
\begin{figure}[!t]
\centering
\includegraphics[width=0.4\textwidth]{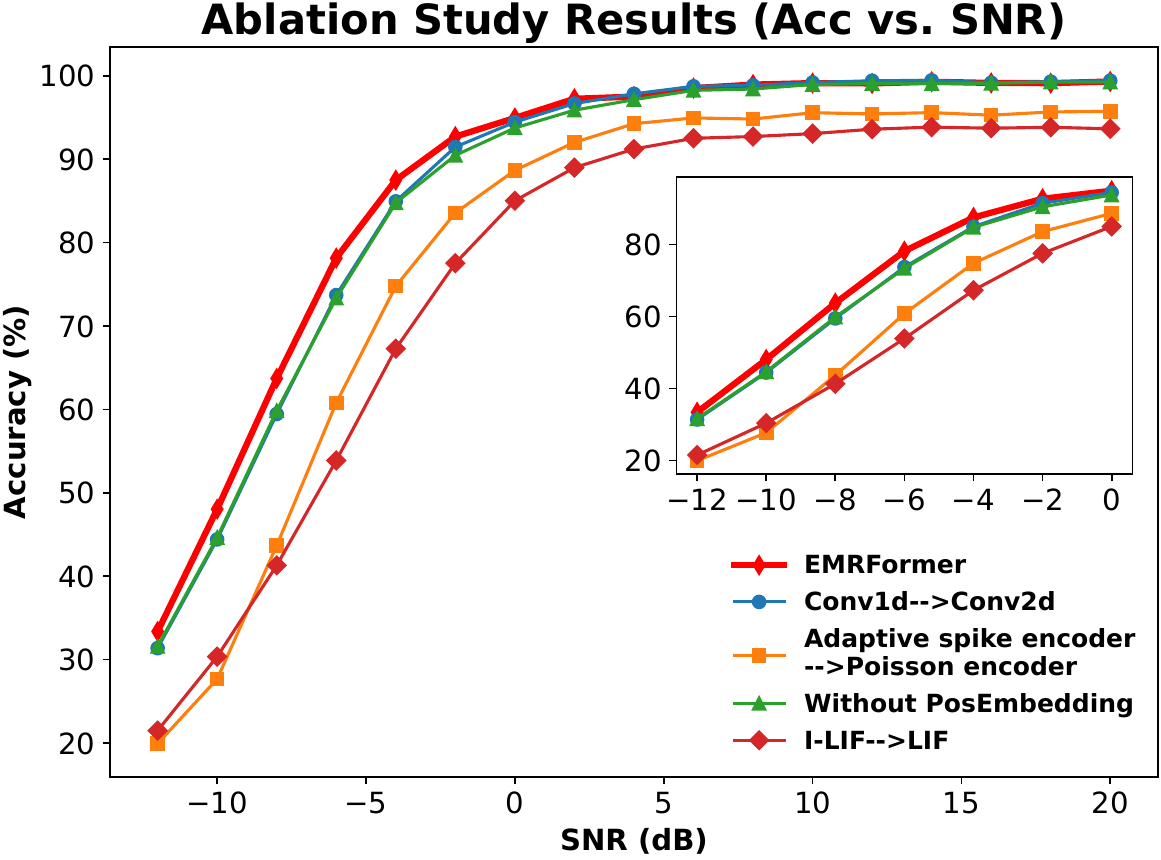}
\caption{Accuracy-SNR curves under different settings}
\label{fig:ablation}
\end{figure}

\subsection{Energy-efficient Inference on Neuromorphic Chips}
We also measure the real-time power of SNNs on neuromorphic chips. We have assessed \modelname's inference speed and Power on the KA200 chip (batch size = 1) and have compared its results with a NVIDIA GeForce RTX 3090 GPU and a Jetson Orin NX. The models are deployed to the GPU and Orin NX via ONNX (compiler), while we utilize Lyngor to conduct hardware mapping on KA200. 
As per Table \ref{tab:KA200}, it demonstrates the Algorithm Power on KA200, outperforming the GPU and Orin NX by 5$\times$. 
These findings confirm the efficient hardware mapping of SNN models on neuromorphic chips and highlight the proposed model's advantages in low-power, low-latency, and high-energy-efficiency contexts.
\begin{table}[]
\centering
\caption{The performance of the \modelname across Hardware}
\label{tab:KA200}
\begin{tabular}{ccccc}
\hline
Device                   & GPU & ORIN NX & KA200 \\ \hline
Static power (W)          & 23.48    & 5.95      & 11.12       \\
Operating power (W)         & 31.35    & 8.06     & 11.48      \\
Dynamic Power (W)         & 7.87    & 2.11      & 0.36        \\
Inference Speed (FPS)        & 884.39    & 186.69      &  1281.65       \\ \hline
\multicolumn{4}{c}{\textit{* Power consumption is monitored using build-in sdk.}} \\
\end{tabular}
\end{table}

\section{Conclusion}


In this paper, we present \modelname, a novel end-to-end SNN framework for AMR tasks of radar and communication signal processing. The mode conducts neuromorphic computing by integrating SNNs into a transformer-based architecture to overcome the common challenge of balancing accuracy and energy consumption. The experiments show that \modelname attains top-tier average accuracy across modulation recognition datasets, akin to existing models in parameter scale, yet reduces the theoretical energy consumption by over 90\%. When deployed on the KA200 neuromorphic chip, it boasts roughly 5 times lower power than GPU and Orin NX platforms. This advancement paves the way for efficient real-time AMR on resource-limited devices. In our future work, we plan to delve into end-to-end training and real-world on-chip validation to bolster the framework's robustness and practicality on more application tasks.

\bibliographystyle{IEEEtran}  
\bibliography{refs}           
\end{document}